%% file: main.tex
\PassOptionsToPackage{table}{xcolor}
\documentclass[sigconf,nonacm]{acmart}

\setcopyright{none}
\makeatletter
\let\hyxmp@parse@acmart\relax
\makeatother
\makeatletter\global\@ACM@balancefalse\makeatother

\usepackage{graphicx}
\usepackage{booktabs}
\usepackage{array}
\usepackage{multirow}
\usepackage{amsmath}
\usepackage{mathtools}
\usepackage{xspace}
\usepackage{placeins}
\usepackage{stfloats}
\usepackage{tikz}
\usetikzlibrary{positioning,arrows.meta,calc,fit,backgrounds,shapes.geometric}
\usepackage{pgfplots}
\pgfplotsset{compat=1.18}
\usepgfplotslibrary{groupplots,fillbetween}
\usepackage[capitalise,noabbrev]{cleveref}

\graphicspath{{images/}{images/teaser_panels/}}

\input{math_commands}

\definecolor{tabfirst}{rgb}{1,0.70,0.70}
\definecolor{tabsecond}{rgb}{1,0.85,0.70}
\definecolor{tabthird}{rgb}{1,1,0.70}
\newcommand{\best}[1]{\textbf{#1}}
\newcommand{\s}[1]{\textcolor{gray}{\scriptsize{$\pm #1$}}}

\newcommand{\ours}{SG-t-SNE-$\Pi$\xspace}

\newcommand{\lambdalens}{LambdaLens\xspace}

\newcommand{\pbmc}{PBMC-8k\xspace}
\newcommand{\LTh}{\mathrm{LT}}
\newcommand{\LCh}{\mathrm{LC}}
\newcommand{\LTCh}{\mathrm{LTC}_{h}}
\newcommand{\Proc}{\mathrm{Proc}}

\input{images/_figstyle}

\theoremstyle{definition}
\newtheorem{definition}{Definition}
\theoremstyle{plain}
\newtheorem{proposition}{Proposition}
\newtheorem{corollary}[proposition]{Corollary}
\theoremstyle{remark}

\crefname{definition}{Definition}{Definitions}
\Crefname{definition}{Definition}{Definitions}
\crefname{proposition}{Proposition}{Propositions}
\Crefname{proposition}{Proposition}{Propositions}
\crefname{corollary}{Corollary}{Corollaries}
\Crefname{corollary}{Corollary}{Corollaries}
\crefname{remark}{Remark}{Remarks}
\Crefname{remark}{Remark}{Remarks}

\title{LambdaLens: Auditing a Representation-Changing Hyperparameter in Graph Neighbor Embedding}

\author{Juntang Wang}
\affiliation{%
  \institution{Harvard University}
  \city{Cambridge}\country{USA}}
\email{juntangwang@g.harvard.edu}

\author{Yihan Wang}
\affiliation{%
  \institution{Duke Kunshan University}
  \city{Kunshan}\country{China}}
\email{yihan.wang@dukekunshan.edu.cn}

\author{Sheng'en Li}
\affiliation{%
  \institution{Duke Kunshan University}
  \city{Kunshan}\country{China}}
\email{shengen.li@dukekunshan.edu.cn}

\author{Siyuan Cao}
\affiliation{%
  \institution{Duke Kunshan University}
  \city{Kunshan}\country{China}}
\email{siyuan.cao@dukekunshan.edu.cn}

\author{Liming Zhu}
\affiliation{%
  \institution{Duke Kunshan University}
  \city{Kunshan}\country{China}}
\email{liming.zhu@dukekunshan.edu.cn}

\author{Man Zhang}
\affiliation{%
  \institution{Westlake University}
  \city{Hangzhou}\country{China}}
\email{zhangman@westlake.edu.cn}

\author{Dongmian Zou}
\authornote{Corresponding authors.}
\affiliation{%
  \institution{Duke Kunshan University}
  \city{Kunshan}\country{China}}
\email{dongmian.zou@dukekunshan.edu.cn}

\author{Shixin Xu}
\authornotemark[1]
\affiliation{%
  \institution{Duke Kunshan University}
  \city{Kunshan}\country{China}}
\email{shixin.xu@dukekunshan.edu.cn}

\renewcommand{\shortauthors}{Juntang Wang et al.}

\begin{abstract}
Changing a graph embedding parameter can change what the layout represents. In \ours, the rescaling parameter $\lambda$ changes the probability assigned to each neighbor before optimization begins. We introduce \lambdalens to trace that change from individual edges to the resulting layout. For nodes with at least two neighbors, we derive a normalized power path. When the original weights are unequal, the number of neighbors determines where the weights become uniform and where their ordering reverses. Near $\lambda=1$, the rate of change in each edge probability depends on its surprisal relative to the neighborhood entropy. To measure the geometric effect, we vary $\lambda$ while holding the input graph, initialization, random seed, and optimizer settings fixed. Across six graphs, some layouts continue to move as $\lambda$ increases, others stabilize early, and others barely move. PubMed has the widest relative spread in node degree, yet its layout changes little. On four datasets with class labels, four probes recover $98.48$--$100\%$ of the best label-fidelity score on a 16-value grid, using $75\%$ fewer evaluations. In a six-dataset comparison, \ours attains the highest observed primary score on four datasets and ranks among the top three on all six. On Citeseer, four probes come close to the dense-grid optimum, yet PHATE better preserves class structure, making a change of embedding pipeline worth considering on that graph.
\end{abstract}



\hypersetup{pdfauthor={Juntang Wang, Yihan Wang, Sheng'en Li, Siyuan Cao, Liming Zhu, Man Zhang, Dongmian Zou, Shixin Xu}}
\begin{document}

\maketitle

%% =================================================================
%% Introduction
%% =================================================================
\section{Introduction}
\label{sec:introduction}

If a cluster splits after a change in parameter setting, did the optimizer rearrange the same neighborhoods, or did the neighborhoods themselves change? Both explanations are plausible when a parameter acts before optimization. In t-SNE, perplexity changes the probability assigned to each neighbor. In a graph layout, $k$ can change which edges exist. In a diffusion embedding, diffusion time changes the operator that defines distance. In each case, comparing the layouts requires knowing what the optimizer was asked to preserve.

A parameter sweep shows where points move. Interpreting that movement requires tracing the parameter back to the input representation. Which relationships change, and does the same setting have the same effect at every node? Once the target is known, two further questions follow. How much does the layout move when the initialization is held fixed? Does that movement preserve more of the structure relevant to the task?

We study the rescaling parameter $\lambda$ in \ours~\cite{pitsianis2019hpec,pitsianis2019joss}. This method takes a sparse stochastic graph as input and directly optimizes the t-SNE objective. Before optimization, it transforms each column of the graph, which contains the probabilities assigned to one node's neighbors. The original method specifies the columnwise transformation. We derive what it does to the relative weights of edges and trace its effect through the embedding.

Every node with at least two neighbors follows its own \emph{normalized power path}. For node $j$, \ours chooses an exponent $\gamma_j(\lambda)$ such that $\sum_i p_{i\mid j}^{\gamma_j}=\lambda$, then normalizes the column. The ratio of two neighbor probabilities becomes $(p_{i\mid j}/p_{k\mid j})^{\gamma_j}$. Let $s_j$ be the number of neighbors. For unequal affinities, $\lambda<1$ sharpens their differences, $1<\lambda<s_j$ flattens them, $\lambda=s_j$ gives all neighbors equal weight, and $\lambda>s_j$ reverses their ordering. A single value of $\lambda$ can thus flatten the affinities at one node and reverse them at another. Near the identity, $\lambda=1$, the rate of change also depends on the initial probabilities. A neighborhood with equal affinities stays uniform throughout the sweep.

\lambdalens pairs embeddings so that these changes can be followed from the graph to the layout. For each graph and seed, all values of $\lambda$ share the same starting coordinates, optimizer settings, and evaluation nodes. We align the layouts, measure how far the nodes move, and score the structure each layout preserves. The resulting curves show where a graph responds to rescaling and where further changes have little effect.

We make four contributions.
\begin{itemize}
  \item \textbf{Local edge weights.} We derive the four phases of the normalized power path, characterize neighborhood entropy along it, and obtain the first-order change in each edge probability at $\lambda=1$.
  \item \textbf{Paired measurements.} We link the transformed target to aligned node movement and task fidelity while controlling graph construction, preprocessing, random seed, initialization, optimizer settings, evaluation subset, and visual encoding.
  \item \textbf{Graph-specific response.} Across six graphs, we find extended, saturating, and nearly inert responses. PubMed shows why degree dispersion alone is a poor guide to layout movement. Citeseer illustrates the difference between improving the selected setting and changing the embedding pipeline.
  \item \textbf{Search and comparison.} On four labeled datasets, four probes retain $98.48$--$100\%$ of the best 16-point label-based objective. In the separate six-dataset benchmark, \ours ranks first on four datasets and among the top three on all six.
\end{itemize}

\section{Related Work}
\label{sec:related}

\subsection{Neighbor embedding and graph-native layout}

Classical t-SNE converts pairwise distances into local conditional probabilities and fits a low-dimensional Student-$t$ model by minimizing KL divergence~\cite{vandermaaten2008tsne}. Barnes--Hut and interpolation methods accelerate this optimization~\cite{vandermaaten2014bhtsne,linderman2019fitsne}. UMAP constructs a fuzzy neighborhood graph and optimizes a cross-entropy objective~\cite{mcinnes2018umap,mcinnes2018umapjoss}. LargeVis~\cite{tang2016largevis}, ForceAtlas2~\cite{jacomy2014forceatlas2}, and DRGraph~\cite{zhu2021drgraph} also use graph structure to visualize large datasets. Node2vec learns a representation from random walks before a second projection stage~\cite{grover2016node2vec}. Recent work connects dimensionality reduction to graph drawing~\cite{paulovich2025drmeetsgraph} and relates neighbor embedding to contrastive and graph-based objectives~\cite{bohm2023tsimcne,damrich2023contrastive,bohm2025nodeembeddings}.

\ours accepts a sparse stochastic graph directly and retains the probability-matching objective of t-SNE~\cite{pitsianis2019hpec,pitsianis2019joss}. Its rescaling step generates a family of target graphs, one for each value of $\lambda$. This lets us examine how the edge weights change before measuring how far their corresponding layouts move.

\subsection{Hyperparameter effects, instability, and visual comparison}

Wattenberg et al.~\cite{wattenberg2016tsne} and Kobak and Berens~\cite{kobak2019art} describe how t-SNE settings affect the interpretation of a layout. Initialization itself can preserve global organization or introduce variation~\cite{kobak2021initialization}. GhostUMAP measures pointwise instability across repeated UMAP runs~\cite{jung2024ghostumap}, and scDEED detects dubious local structures in single-cell embeddings~\cite{xia2024scdeed}. HyperNP supports visual exploration of hyperparameter effects~\cite{appleby2022hypernp}, while feature-level explanations identify attributes associated with changes across projections~\cite{ovcharenko2024featureclock}. Parametric UMAP learns an out-of-sample mapping~\cite{sainburg2021parametricumap}.

Our question concerns the source of a layout change. We start with the change in high-dimensional neighbor probabilities, measure node movement after aligning the layouts, and score fidelity to the task. Pairing the initialization within each seed lets us compare values of $\lambda$ from a common starting point.

\subsection{Embedding evaluation and auditability}

Trustworthiness and continuity quantify which neighborhoods an embedding creates or breaks~\cite{venna2001trustworthiness}. Procrustes analysis compares geometries after alignment~\cite{gower1975procrustes}. The choice of metric matters because embedding quality depends on the data and the task~\cite{espadoto2021dr,cashman2025dr_critical}. For labeled data, Label-Trustworthiness and Label-Continuity measure whether an embedding introduces or destroys class structure while allowing a class to occupy several clusters~\cite{jeon2023labeltnc}. ZADU implements these and other reliability measures~\cite{jeon2023zadu}.

We use the harmonic mean of Label-Trustworthiness ($\LTh$) and Label-Continuity ($\LCh$) to choose among candidate values of $\lambda$. We denote this score by $\LTCh$. In the cross-method comparison, we report $\LTh$ for labeled data and conventional trustworthiness for the unlabeled graph.

\section{Preliminaries}
\label{sec:preliminaries}

\subsection{Sparse stochastic graph and preprocessing}

Let $G=(V,E,A)$ be a sparse graph with $n=|V|$. We use a column-stochastic convention: for node $j$, $p^{(j)}=(p_{i\mid j})_{i\in S_j}$ is a strictly positive probability vector on support $S_j=\{i:(i,j)\in E\}$, $\sum_{i\in S_j}p_{i\mid j}=1$, and $s_j=|S_j|\ge 2$. A row-stochastic implementation is equivalent after transposition. We give the treatment of isolated nodes and nodes with a single neighbor in \cref{app:support}.

The citation, MNIST-kNN, and ca-AstroPh graphs have symmetric, unweighted edges. We weight each edge by the Jaccard similarity of its endpoints' neighborhoods and normalize each column to sum to one (\texttt{unweighted\_to\_weighted}). Edges can then differ in weight even when they were identical in the binary adjacency. The \pbmc graph already has stochastic weights ($k=30$), so we use it directly (\texttt{unweighted\_to\_weighted=False}). This choice matters because a column of equal weights stays uniform for every $\lambda$ (\cref{cor:uniform}).

\subsection[The SG-t-SNE-Pi target and objective]{The \ours target and objective}
\label{sec:sgtsne_prelim}

The original \ours transformation solves one exponent per neighborhood~\cite{pitsianis2019hpec}. For a global $\lambda>0$, define
\begin{equation}
  Z_j(\gamma)=\sum_{i\in S_j}p_{i\mid j}^{\gamma},
  \qquad
  Z_j\!\left(\gamma_j(\lambda)\right)=\lambda .
  \label{eq:partition}
\end{equation}
The normalized rescaled conditional distribution is
\begin{equation}
  \widetilde p_{i\mid j}^{(\lambda)}
  =\frac{p_{i\mid j}^{\gamma_j(\lambda)}}{\lambda},
  \qquad i\in S_j.
  \label{eq:normalized_path}
\end{equation}
Symmetrization produces the high-dimensional target
\begin{equation}
  P_{ij}^{(\lambda)}
  =\frac{\widetilde p_{i\mid j}^{(\lambda)}+
          \widetilde p_{j\mid i}^{(\lambda)}}{2n},
  \qquad P_{ii}^{(\lambda)}=0,
  \label{eq:sym_target}
\end{equation}
where absent directed edges contribute zero. Because every normalized column sums to one, $\sum_{i\ne j}P_{ij}^{(\lambda)}=1$.

For an embedding $Y=(y_1,\ldots,y_n)^\top\in\mathbb{R}^{n\times 2}$, \ours uses Student-$t$ affinities
\begin{equation}
  Q_{ij}(Y)=
  \frac{(1+\lVert y_i-y_j\rVert_2^2)^{-1}}
       {\sum_{k\ne \ell}(1+\lVert y_k-y_\ell\rVert_2^2)^{-1}}
  \label{eq:low_affinity}
\end{equation}
and minimizes $\mathrm{KL}(P^{(\lambda)}\Vert Q(Y))$. Changing $\lambda$ therefore changes the target distribution before the optimizer takes its first step.

\subsection{Evaluation quantities and notation}
\label{sec:metrics_prelim}

For labeled data, $\LTh$ and $\LCh$ denote Label-Trustworthiness and Label-Continuity~\cite{jeon2023labeltnc}. We use
\begin{equation}
  \LTCh(Y)=\frac{2\LTh(Y)\LCh(Y)}{\LTh(Y)+\LCh(Y)}
  \label{eq:ltch}
\end{equation}
for the four-probe search and to annotate the layouts. Since $\LCh$ is frequently close to its ceiling, we report raw $\LTh$ in the cross-method comparison. On the unlabeled ca-AstroPh graph, we use conventional trustworthiness $T$ at $k=15$~\cite{venna2001trustworthiness}. We compute metrics with ZADU~\cite{jeon2023zadu}. For datasets larger than $5{,}000$ nodes, all settings at a given seed share an evaluation subsample of at most $5{,}000$ nodes.

To compare two embeddings of the same nodes, we center both and solve an orthogonal Procrustes alignment, allowing rotation and reflection while preserving scale. The normalized distance used in the diagnostic is
\begin{equation}
  \Proc(Y,Y')=
  \frac{\min_{R^\top R=I}\sqrt{n^{-1}}\lVert Y_c-Y'_cR\rVert_F}
       {\sqrt{n^{-1}}\lVert Y_c\rVert_F}.
  \label{eq:procrustes}
\end{equation}
The denominator is the root-mean-square distance from the center of the reference layout $Y$. Because this makes the distance directional, we use a consistent ordering of layouts in each pairwise summary.

\section{The LambdaLens Audit}
\label{sec:methodology}

\subsection{The normalized local power path}

\begin{definition}[Local $\lambda$ path]
For a stochastic neighborhood $p^{(j)}$ with support size $s_j\ge2$, the \emph{local $\lambda$ path} is the family $\widetilde p^{(j,\lambda)}$ in \cref{eq:normalized_path}, where $\gamma_j(\lambda)$ is the unique solution of \cref{eq:partition}.
\end{definition}

Raising each entry to the local exponent gives the column total mass $\lambda$. Dividing by that mass produces the probabilities used in \cref{eq:sym_target}. Their ratios tell us which neighbors gain weight relative to others.

\begin{proposition}[Sharpen--flatten--uniformize--invert phases]
\label{prop:phases}
For every $\lambda>0$, $\gamma_j(\lambda)$ exists and is unique. Moreover,
\begin{equation}
\gamma_j(\lambda)=
\begin{cases}
>1, & 0<\lambda<1,\\
=1, & \lambda=1,\\
\in(0,1), & 1<\lambda<s_j,\\
=0, & \lambda=s_j,\\
<0, & \lambda>s_j.
\end{cases}
\label{eq:phase_cases}
\end{equation}
For any $i,k\in S_j$,
\begin{equation}
  \frac{\widetilde p_{i\mid j}^{(\lambda)}}
       {\widetilde p_{k\mid j}^{(\lambda)}}
  =\left(\frac{p_{i\mid j}}{p_{k\mid j}}\right)^{\gamma_j(\lambda)}.
  \label{eq:ratio_path}
\end{equation}
Thus $\lambda<1$ sharpens relative affinities, $1<\lambda<s_j$ flattens them without changing order, $\lambda=s_j$ makes the neighborhood uniform, and $\lambda>s_j$ reverses every strict ordering.
\end{proposition}

\noindent\textit{Proof sketch.}
Every factor $p_{i\mid j}^{\gamma}\log p_{i\mid j}$ is negative, so $Z_j$ is strictly decreasing. Together with $Z_j(1)=1$, $Z_j(0)=s_j$, and its limits at $\pm\infty$, this gives existence, uniqueness, and the cases in \cref{eq:phase_cases}. Cancelling the common normalization factor gives \cref{eq:ratio_path}. The full proof is in \cref{app:proofs}.

The boundary in \cref{prop:phases} depends on the node. If $s_j<\lambda<s_k$, the ordering at node $j$ has reversed while node $k$ is still in the flattening phase. The rate of change also depends on how uneven the original weights are.

\begin{proposition}[Entropy and identity sensitivity]
\label{prop:entropy}
Let $H_j(\lambda)=-\sum_i\widetilde p_{i\mid j}^{(\lambda)}\log \widetilde p_{i\mid j}^{(\lambda)}$. Then
\begin{equation}
 \frac{dH_j}{d\lambda}
 =-\frac{\gamma_j(\lambda)\,
 \mathrm{Var}_{\widetilde p^{(j,\lambda)}}(\log p_{i\mid j})}
 {\lambda\,
 \mathbb{E}_{\widetilde p^{(j,\lambda)}}[\log p_{i\mid j}]}.
 \label{eq:entropy_derivative}
\end{equation}
For a nonuniform neighborhood, entropy increases for $\lambda<s_j$ and decreases after the order-inversion boundary $\lambda>s_j$. At the identity,
\begin{equation}
 \left.\frac{d\widetilde p_{i\mid j}^{(\lambda)}}{d\lambda}\right|_{\lambda=1}
 =p_{i\mid j}\,
 \frac{-\log p_{i\mid j}-H(p^{(j)})}{H(p^{(j)})}.
 \label{eq:identity_derivative}
\end{equation}
\end{proposition}

In \cref{eq:identity_derivative}, the surprisal of edge $i$ is $-\log p_{i\mid j}$ and its mean over the neighborhood is the entropy $H(p^{(j)})$. As $\lambda$ increases from one, an edge gains probability if its surprisal exceeds this mean and loses probability if it falls below it. The support size determines where all weights become equal. To determine how fast an individual weight changes, we also need the initial probability distribution.

\begin{corollary}[Uniform-neighborhood invariance]
\label{cor:uniform}
If $p_{i\mid j}=1/s_j$ for all $i\in S_j$, then $\widetilde p_{i\mid j}^{(\lambda)}=1/s_j$ for every $\lambda>0$.
\end{corollary}

On a binary graph, degree normalization alone gives equal weight to every neighbor. By \cref{cor:uniform}, changing $\lambda$ would then leave every column unchanged. Jaccard weighting lets rescaling act wherever neighborhood overlap yields unequal affinities.

\subsection{Paired attribution protocol}
\label{sec:paired_protocol}

For dataset $D$, seed $r$, and setting $\lambda$, \lambdalens computes the layout
\begin{equation}
  Y_{D,r}^{(\lambda)}=
  \mathcal{O}\!\left(P_D^{(\lambda)};
  Y_{0,D,r},\theta_D,r\right),
  \label{eq:paired_run}
\end{equation}
where $\mathcal{O}$ is the \ours optimizer and $\theta_D$ contains its non-$\lambda$ settings. At each seed, all runs start from the same adjacency-SVD coordinates $Y_{0,D,r}$ and use the same evaluation nodes. Within a pair, only the target changes. Repeating the comparison over seeds measures how the scores vary with the seed.

We compare the runs at three levels (\cref{fig:audit_workflow}).
\textbf{Target.} For each graph, we record the preprocessing, whether edges receive Jaccard weights, the support size $s_j$ of each neighborhood, and the transformation applied to its probabilities.
\textbf{Geometry.} We use the same visual encoding for all layouts in a sweep and measure aligned node movement with \cref{eq:procrustes}.
\textbf{Fidelity.} On the shared evaluation nodes we compute $\LTh$, $\LCh$, $\LTCh$, or $T$ as appropriate. Reading the score alongside the movement reveals when visibly different layouts preserve class or neighborhood structure equally well.

\input{images/02_audit_workflow}

\subsection{Empirical response fingerprints}
\label{sec:fingerprints}

We use three terms to describe the response over the tested grid. In an \emph{extended response}, nodes move across several adjacent $\lambda$ intervals and fidelity changes with them. In a \emph{saturating response}, most movement or fidelity gain occurs in the first few intervals, followed by a plateau. In a \emph{nearly inert response}, the layout and its fidelity score both remain close to their starting values. These descriptions refer to the observed curves on a given graph and grid.

\subsection{Efficient label-assisted probe search}
\label{sec:probe_search}

For labeled datasets we also test whether a small probe set can retain the objective of a denser search. Let
\begin{align}
 \Lambda_{4}&=\{1,5,20,50\},\\
 \Lambda_{16}&=\{0.1,0.2,0.5,1,2,3,5,7,10,15,20,30,50,80,100,150\}.
\end{align}
At seed $42$ with the paired adjacency-SVD initialization, we compute
\begin{equation}
 \widehat\lambda_{m}
 =\arg\max_{\lambda\in\Lambda_m}\LTCh\!\left(Y^{(\lambda)}\right),
 \qquad m\in\{4,16\}.
 \label{eq:probe_search}
\end{equation}
We divide the best four-probe score by the best dense-grid score to measure objective retention.

\section{Experiments}
\label{sec:experiments}

We first vary $\lambda$ within each graph to measure how much the layout moves and how its fidelity changes. We then measure how much label fidelity four probes recover on a 16-point grid. A separate benchmark compares \ours with graph- and feature-based alternatives.

\input{images/t00_datasets}

\subsection{Datasets and provenance}

\Cref{tab:datasets} summarizes the six datasets and their inputs. We load the Planetoid citation graphs and labels in node-ID order~\cite{sen2008cora}. For MNIST we build an unweighted 15-nearest-neighbor graph on all $70{,}000$ images~\cite{lecun1998mnist}, using scikit-learn's default Minkowski distance with $p=2$ (Euclidean distance)~\cite{pedregosa2011sklearn}. ca-AstroPh is an unlabeled coauthorship graph from SNAP~\cite{leskovec2014snap}. For \pbmc we use the stochastic $k$NN graph distributed with the original \ours example~\cite{pitsianis2019hpec}. To obtain a reference partition independent of $\lambda$, we apply Ward agglomerative clustering with $k=7$ to the top 15 nontrivial normalized-Laplacian eigenvectors. The resulting pseudo-labels measure agreement with this spectral partition. Biological cell-type annotations are unavailable for this graph.

\subsection{Paired response protocol}

We evaluate all 16 values in $\Lambda_{16}$ at seed $42$. To initialize the embedding, we compute two components of the adjacency matrix with sparse \texttt{TruncatedSVD} (\texttt{random\_state=seed}), center the coordinates, and rescale them to $\max|Y_0|=10^{-4}$. All $\lambda$ values in a dataset/seed block share this $Y_0$. The layout portraits show four settings ($1$, $5$, $20$, and $80$). We average the normalized Procrustes distance over all pairs of settings on the full 16-point grid and compare it with the coefficient of variation in node degree, $\mathrm{CV}(d)=\mathrm{sd}(d)/\mathrm{mean}(d)$.

With random initialization, the normalized Procrustes variation across seeds was roughly $17$--$27\%$ on the tested graphs. Reusing the initial coordinates within each sweep removes this source of variation from the comparison across $\lambda$.

\subsection{Probe-search protocol}

We compare the two grids in \cref{eq:probe_search} on the four datasets with external class labels (Cora, Citeseer, PubMed, and MNIST-kNN). At seed $42$ the searches use the same preprocessing, initial coordinates, and metric. We report the selected $\lambda$, its $\LTCh$, and the relative loss against the 16-point optimum.

\subsection{Comparative benchmark protocol}
\label{sec:baseline_protocol}

We compare the audited pipeline with four representative baselines under fixed, reproducible configurations. The baselines are UMAP~\cite{mcinnes2018umap}, openTSNE~\cite{policar2024opentsne}, PHATE~\cite{moon2019phate}, and node2vec followed by UMAP~\cite{grover2016node2vec}. UMAP uses $n_{\mathrm{neighbors}}=15$, $\mathrm{min\_dist}=0.1$, and 1,000 epochs. openTSNE uses perplexity 30 and 1,000 iterations. PHATE uses $k=15$. Node2vec uses 64 dimensions, walk length 30, 200 walks, and $p=q=1$, followed by the same UMAP head. Methods that accept initial coordinates use a two-dimensional adjacency-SVD layout. The baseline runs on PubMed, \pbmc, and ca-AstroPh reuse one initialization per graph across seeds.

Each method uses its native input. When features are available, UMAP, openTSNE, and PHATE take the feature matrix, while \ours and node2vec use the graph. On graph-only datasets, UMAP takes the precomputed distance $1-A_{\mathrm{sym}}$ and both openTSNE and PHATE take precomputed affinities. PHATE runs on subsamples of $10{,}000$ nodes for MNIST and $3{,}000$ for ca-AstroPh. For node2vec+UMAP on ca-AstroPh, five seeds contribute to the quality estimate, and one run meets the pinned-thread timing protocol. On MNIST, node2vec+UMAP did not finish within the compute budget.

In the benchmark, the four-probe search supplies the \ours settings for the citation graphs, giving $\lambda=20$ on Cora, $1$ on Citeseer, and $5$ on PubMed. The benchmark uses fixed settings of $20$ on MNIST-kNN and $10$ on \pbmc and ca-AstroPh. The four-probe result for MNIST in \cref{tab:probe_search} belongs to the search experiment.

We use seeds 42--46 ($N=5$) on Cora, PubMed, and ca-AstroPh and seeds 42--51 ($N=10$) on Citeseer, MNIST-kNN, and \pbmc. Node2vec+UMAP contributes five seeds on Citeseer and \pbmc. Quality estimates for \ours on PubMed, MNIST-kNN, \pbmc, and ca-AstroPh, and for UMAP on MNIST-kNN, come from Linux runs initialized separately at each seed. Wall-clock times come from Apple M3 Pro runs with eight process workers and \texttt{OMP\_NUM\_THREADS=1}. \Cref{app:stats} specifies the paired comparisons.

\subsection[Ablation and non-lambda sensitivity]{Ablation and non-$\lambda$ sensitivity}

For the ablation, the full configuration uses $\lambda=20$ on Cora and $5$ on \pbmc. We measure the effect of each change to \ours using raw $\LTh$ over seeds 42--46. We omit adjacency-SVD initialization, omit Jaccard weighting where applicable, or fix $\lambda$ at $1$ or $20$. In separate sweeps over seeds 42--44, we vary the perplexity used to construct an upstream openTSNE affinity, the iteration count, the MNIST neighborhood size $k$, and the early-exaggeration factor $\alpha$. The $k$ sweep uses a 10,000-image cosine-kNN graph. The primary MNIST comparison uses the full 70,000-node Euclidean graph.

\section{Results}
\label{sec:results}

\subsection[Graph-specific response fingerprints]{Graph-specific response fingerprints}

Cora continues to change across the four settings shown in \cref{fig:response_portrait}. Classes separate and reconnect as $\lambda$ increases. The five-seed fidelity summary rises from $0.922$ at $\lambda=1$ to $0.966$ at $20$, then declines. This is an extended response. On \pbmc, most of the change occurs between $1$ and $5$. Both the layout and its agreement with the spectral pseudo-labels then stabilize. MNIST-kNN is nearly inert over the displayed settings. Its layouts look similar, and $\LTCh$ varies by a few thousandths.

\input{images/01_response_portrait}

The local power path helps explain why uniformly increasing $\lambda$ can produce different layout responses. A neighborhood flattens until $\lambda=s_j$. Beyond that point its strongest and weakest affinities trade places. Nodes differ in both support size and initial probabilities, so the same graph can contain neighborhoods in several phases at once. Cora's sustained movement is consistent with substantial changes to its weighted neighborhoods. On MNIST, Jaccard weighting also introduces unequal affinities, yet rescaling them leaves the overall digit layout stable over the displayed range.

PubMed illustrates why degree dispersion alone gives little guidance about how far a layout will move (\cref{fig:cv_procrustes}). It has the highest degree coefficient of variation ($1.653$), but its mean pairwise movement is $0.193$, compared with $0.854$ on Cora and $0.760$ on Citeseer. MNIST has both low degree dispersion and low movement ($0.095$). ca-AstroPh falls between the extremes, while \pbmc has moderate degree dispersion and low movement. Degree counts describe how many edges each node has. The derivative in \cref{eq:identity_derivative} also depends on how probability is distributed among those edges, and the layout depends in turn on the optimizer's response to the resulting target.

\input{images/04_diagnostic}

\subsection{Four probes retain the dense-grid objective}

\input{images/t01_probe_search}

Both searches select $\lambda=5$ on PubMed (\cref{tab:probe_search}). On Cora and MNIST-kNN, the dense grid finds optima outside the four probes, at $7$ and $30$, respectively. Restricting the search to four settings reduces the score by about $0.04\%$ on Cora and $0.13\%$ on MNIST-kNN. Citeseer shows the largest loss ($1.52\%$). Four probes select $\lambda=1$ with a score of $0.8536$, while the dense grid selects $\lambda=3$ with a score of $0.8668$.

\subsection{Fidelity and runtime across methods}

\input{images/t02_comparison}

\input{images/03_sensitivity}

\ours records the highest primary score on Cora, PubMed, \pbmc, and ca-AstroPh and ranks among the top three methods with scores on all six datasets (\cref{tab:comparison}).

\textbf{Citation and cell graphs.}
On Cora, \ours reaches $\LTh=0.932\pm0.013$, ahead of node2vec+UMAP ($0.915\pm0.007$) on four of five seeds. Their M3 Pro runs take 11 and 420 seconds, respectively. On PubMed, \ours scores higher than PHATE on all five seeds ($0.905\pm0.005$ versus $0.895\pm0.008$). We report both five-seed comparisons descriptively.

On \pbmc, \ours scores $0.989\pm0.003$ against the spectral pseudo-labels, compared with $0.981\pm0.011$ for UMAP. It scores higher on nine of ten seeds, with a Hodges--Lehmann paired difference of $+0.004$ (exact interval $[0.003,0.019]$). The unadjusted and within-family Holm $p$ values are both $0.049$. Bonferroni adjustment over all four comparisons gives $p=0.195$. This score measures agreement with the spectral partition, and the ten UMAP runs share one initialization. Label-free graph trustworthiness favors \ours on all ten seeds ($0.705\pm0.002$ versus $0.681\pm0.003$, within-family Holm $p=0.004$).

On the unlabeled coauthorship graph ca-AstroPh, \ours and node2vec+UMAP reach $T=0.622\pm0.003$ and $0.621\pm0.003$, respectively, over five seeds. Their wall-clock times are 30 and 4,083 seconds. The latter comes from the single node2vec+UMAP run with pinned threads.

\textbf{Citeseer and MNIST-kNN.}
PHATE reaches $\LTh=0.855\pm0.006$ on Citeseer, compared with $0.754\pm0.012$ for \ours. It scores higher on all ten seeds, with a mean paired difference of $0.101$ (95\% interval $[0.093,0.110]$, within-family Holm $p=0.004$). Four probes recover about $98.5\%$ of the best observed selection score on this graph. This combination makes another embedding pipeline worth testing when a finer grid offers little gain. On MNIST-kNN, UMAP scores $1.000$ and \ours scores $0.991\pm0.007$. UMAP scores higher on all ten seeds, with a mean difference of $0.009$ ($[0.004,0.014]$, within-family Holm $p=0.004$).

\subsection{Ablation and sensitivity}

On Cora, fixing $\lambda=1$ gives the largest drop in $\LTh$ among the ablations. Omitting adjacency-SVD initialization lowers the mean and increases variability (\cref{tab:ablation,app:ablation}). Omitting Jaccard weighting raises Cora's score on three of five seeds and lowers it on the other two. Without this weighting, neighbors have equal affinities and rescaling leaves the target unchanged (\cref{cor:uniform}). A layout can therefore score well even when varying $\lambda$ has no effect.

At $1{,}000$ iterations, all three graphs are close to a plateau in $\LTCh$ (\cref{fig:sensitivity}). In the smaller MNIST cosine-kNN experiment, $k=15$ and $k=30$ give nearly identical peak scores. Early exaggeration at $\alpha=12$ gives the highest score on Cora and \pbmc, while MNIST scores slightly higher at $20$. The largest change in panel (a) comes from reconstructing Cora's input affinities at perplexity 120.

\section{Discussion}
\label{sec:discussion}

\subsection{When the choice of representation matters}

One way to interpret a tuning curve is to ask what choices are available along it. In \ours, $\lambda$ reweights existing edges. Equation~\eqref{eq:ratio_path} shows exactly how far their relative probabilities can change at a given node. On Cora, the identity ablation and the layout portraits show that rescaling improves class fidelity over the identity setting. On Citeseer, a finer grid yields little additional label fidelity and PHATE obtains the higher benchmark score. This makes PHATE's diffusion geometry a candidate for further comparison on Citeseer. The benchmark also differs in native inputs and objectives, so its scores describe complete pipelines.

\subsection{Designing a paired comparison}

Before comparing layouts, inspect the probabilities that reach the optimizer. On a binary graph, omitting the Jaccard conversion gives each neighbor equal weight and leaves the target invariant throughout the sweep. Once the target changes, a shared initialization and evaluation subset allow node movement and fidelity to be compared at successive $\lambda$ values. The shape of the curve matters even when two grids return similar maxima. A plateau indicates that several settings work equally well, whereas an extended response identifies intervals in which the choice still affects the layout or the score.

\subsection{Extending the audit}

The power-path derivation applies to $\phi(x)=x$ in the \ours rescaling rule. It also identifies quantities that could explain individual nodes' responses, including $\gamma_j$, the change in neighborhood entropy, and the divergence between successive target matrices $P^{(\lambda)}$. These quantities could distinguish a change concentrated at a few nodes from one spread across the graph.

The paired design also applies when another parameter changes the graph, kernel, probability distribution, or operator before embedding. Adapting it to dynamic graphs, higher-dimensional embeddings, or judgments by human observers requires specifying which objects can be aligned across settings and which fidelity measure answers the task question.

\section{Conclusion}
\label{sec:conclusion}

A global rescaling parameter has a local meaning. In \ours, a neighborhood with at least two neighbors follows a power path. If its edge weights differ, its support size sets the value of $\lambda$ at which the weights become uniform. Its initial probabilities determine which edges gain weight as $\lambda$ rises from one. \lambdalens follows those changes into paired layouts. Cora keeps moving across multiple settings, \pbmc settles early, and MNIST changes little over the displayed range. On the four datasets with external labels, four evaluations recover $98.48$--$100\%$ of the best 16-point score. Together, the local rule and the paired comparisons explain what a change in $\lambda$ can accomplish on a given graph and when a finer grid is likely to add little.

\begin{acks}
We used AI-assisted tools to edit the prose and prepare the workflow diagram.
\end{acks}

\FloatBarrier
\bibliographystyle{ACM-Reference-Format}
\bibliography{ref}

\appendix
\section{Proofs}
\label{app:proofs}

\subsection[Proof of the phase proposition]{Proof of \cref{prop:phases}}

Fix a node $j$ and suppress its subscript. Because $s\ge2$ and $p_i>0$ with $\sum_i p_i=1$, every $p_i\in(0,1)$. Define $Z(\gamma)=\sum_i p_i^\gamma$. It is continuous and
\[
  Z'(\gamma)=\sum_i p_i^\gamma\log p_i<0,
\]
so it is strictly decreasing. Furthermore, $Z(1)=1$, $Z(0)=s$, $Z(\gamma)\to0$ as $\gamma\to+\infty$, and $Z(\gamma)\to+\infty$ as $\gamma\to-\infty$. The intermediate-value theorem and strict monotonicity yield a unique solution $\gamma(\lambda)$ for every $\lambda>0$ and place it in the five intervals stated in \cref{eq:phase_cases}.

Normalization gives $\widetilde p_i=p_i^{\gamma}/Z(\gamma)$ and
\[
  \frac{\widetilde p_i}{\widetilde p_k}
  =\left(\frac{p_i}{p_k}\right)^\gamma.
\]
Positive exponents preserve the ordering, exponents above one move ratios farther from one, exponents in $(0,1)$ move them toward one, exponent zero makes all ratios one, and negative exponents reverse strict orderings.

\subsection[Proof of the entropy proposition]{Proof of \cref{prop:entropy}}

Write $q_i(\gamma)=p_i^\gamma/Z(\gamma)$ and $\mu(\gamma)=\mathbb{E}_{q(\gamma)}[\log p_i]$. Then
\[
 H(q)=-\sum_i q_i\log q_i
 =\log Z(\gamma)-\gamma\mu(\gamma).
\]
The exponential-family identities are
\[
 \frac{d}{d\gamma}\log Z(\gamma)=\mu(\gamma),
 \qquad
 \frac{d\mu}{d\gamma}=\mathrm{Var}_{q(\gamma)}(\log p_i).
\]
Hence
\[
 \frac{dH}{d\gamma}
 =-\gamma\,\mathrm{Var}_{q(\gamma)}(\log p_i).
\]
Because $\lambda=Z(\gamma)$,
\[
 \frac{d\lambda}{d\gamma}=\lambda\mu(\gamma),
 \qquad
 \frac{d\gamma}{d\lambda}=\frac{1}{\lambda\mu(\gamma)}.
\]
Combining these identities gives \cref{eq:entropy_derivative}. For a nonuniform neighborhood the variance is positive, so $\mu(\gamma)<0$ gives $dH/d\lambda$ the sign of $\gamma$. Entropy increases until $\gamma=0$ ($\lambda=s$) and decreases thereafter. For a uniform neighborhood the variance is zero and entropy is constant.

For the identity derivative,
\[
 \frac{dq_i}{d\gamma}=q_i(\log p_i-\mu).
\]
At $\lambda=1$, $\gamma=1$, $q_i=p_i$, and $\mu=\sum_i p_i\log p_i=-H(p)$. Also $d\gamma/d\lambda=-1/H(p)$. Therefore
\[
 \left.\frac{dq_i}{d\lambda}\right|_{\lambda=1}
 =p_i(\log p_i+H(p))\left(-\frac{1}{H(p)}\right)
 =p_i\frac{-\log p_i-H(p)}{H(p)}.
\]

\subsection[Proof of the uniform-neighborhood corollary]{Proof of \cref{cor:uniform}}

If $p_i=1/s$, then $Z(\gamma)=s^{1-\gamma}=\lambda$, so $\gamma=1-\log_s\lambda$. Consequently,
\[
 \widetilde p_i=\frac{s^{-\gamma}}{\lambda}
 =\frac{s^{-1}\lambda}{\lambda}=\frac{1}{s}.
\]
Thus the normalized distribution is independent of $\lambda$.

\section{Implementation and Reproducibility Details}
\label{app:implementation}

\subsection{Initialization, graph construction, and evaluation}

For each graph and seed, \texttt{pca\_init} initializes two coordinates ($d=2$) with sparse \texttt{TruncatedSVD} using \texttt{random\_state=seed}. We center both columns and rescale by the maximum absolute coordinate to obtain a scale of $10^{-4}$. All $\lambda$ values for that graph and seed share the resulting $Y_0$.

For the primary MNIST graph we call \texttt{kneighbors\_graph} in scikit-learn with the default Minkowski metric and $p=2$. The $k$-sensitivity experiment samples 10,000 nodes and uses cosine distance.

ZADU computes conventional trustworthiness and continuity at $k=15$. We use \texttt{cvm="dsc"} for the label metrics. When $n>5{,}000$, we evaluate a seed-fixed subset of at most 5,000 nodes. For graph-only data, we use dense adjacency or stochastic-graph rows as the ambient representation for these nodes.

The artifact identifies the package revision used for unweighted graphs. It includes the Jaccard conversion before stochastic normalization. An earlier PyPI build omitted the conversion, leaving the unweighted target invariant across $\lambda$.

\subsection{Baseline configurations}

\begin{table*}[t]
\centering
\caption{Baseline settings and inputs for the cross-method comparison. Each baseline uses the same hyperparameters across datasets.}
\label{tab:baseline_settings}
\small
\begin{tabular}{p{0.16\linewidth} p{0.32\linewidth} p{0.42\linewidth}}
\toprule
Method & Core settings & Input handling \\
\midrule
UMAP & $n_{\mathrm{neighbors}}=15$, $\mathrm{min\_dist}=0.1$, 1,000 epochs & Feature matrix when available; precomputed $1-A_{\mathrm{sym}}$ on graph-only inputs; adjacency-SVD initialization where accepted. \\
openTSNE & perplexity 30, 1,000 iterations & Feature matrix when available; \texttt{PrecomputedAffinities(adj)} on graph-only inputs; adjacency-SVD initialization. \\
PHATE & $k=15$ & Feature matrix when available; precomputed-affinity path on graph-only inputs; 10,000-node MNIST and 3,000-node ca-AstroPh subsamples. \\
node2vec+UMAP & 64 dimensions, walk length 30, 200 walks, $p=q=1$; UMAP head as above & Graph input. On ca-AstroPh, five seeds for quality and one pinned-thread run for timing. No completed MNIST run. \\
\ours & Default optimizer, Jaccard conversion for unweighted edges, adjacency-SVD initialization & Four-probe $\lambda=20,1,5$ on Cora, Citeseer, and PubMed. Fixed $\lambda=20$ on MNIST-kNN and $10$ on \pbmc and ca-AstroPh. \pbmc supplies its stochastic weights. \\
\bottomrule
\end{tabular}
\end{table*}

Separate one-factor sweeps on Cora and PubMed at seed 42 vary openTSNE perplexity over $\{10,30,50,100\}$, UMAP neighbors over $\{5,15,30,50,100\}$, PHATE $k$ over $\{5,15,30\}$, and node2vec walk length over $\{40,80,120\}$. The primary comparison uses the fixed baseline settings in \cref{tab:baseline_settings}.

\subsection{Sensitivity grids}

We vary upstream-affinity perplexity over 10, 20, 30, 50, 80, and 120, and the iteration count over 500, 1,000, 2,000, 4,000, and 8,000. For the MNIST cosine-kNN graph we use $k=5,10,15,30,50$. For early exaggeration we use $\alpha=1,5,12,20,50$. Each curve reports the mean $\LTCh$ over seeds 42--44 (\cref{eq:ltch}).

\subsection{Subtractive ablation details}
\label{app:ablation}

\input{images/t03_ablation}

\Cref{tab:ablation} reports raw Label-Trustworthiness. Fixing $\lambda=1$ changes Cora's score from $0.935$ to $0.855$ and \pbmc's from $0.987$ to $0.977$. Removing adjacency-SVD initialization lowers the mean and increases variation across seeds. On Cora, omitting Jaccard weighting gives $\LTh=0.940$. The paired differences from the full configuration are $+0.004$, $-0.009$, $-0.017$, $+0.015$, and $+0.036$ for seeds 42--46. With uniformly weighted neighborhoods, varying $\lambda$ leaves every edge probability unchanged (\cref{cor:uniform}).

All ablations ran in one batch on a Linux x86-64 host. The Cora benchmark in \cref{tab:comparison} used the same $\lambda=20$, Jaccard weighting, per-seed adjacency-SVD initialization, and all 2,708 evaluation nodes on the M3 Pro. On Linux the independently run full and fixed-$20$ rows match at every seed. Two M3 Pro runs at seed 42 agree to $10^{-15}$. Individual scores vary across hosts, consistent with the sensitivity of nonconvex optimization to differences in floating-point arithmetic. The mean scores differ by $0.003$ ($0.935\pm0.019$ on Linux and $0.932\pm0.013$ on the M3 Pro).

\subsection{Neighborhood support in the six graphs}
\label{app:support}

\input{images/t04_support}

\Cref{tab:support} counts how many nodes meet the $s_j\ge2$ condition of \cref{prop:phases}. Before normalization, the implementation sets aside nodes with no edges (48 on Citeseer and one on ca-AstroPh). It embeds the remaining graph, then places the isolated nodes at a single fixed point and includes them in evaluation.

A node with one neighbor assigns it probability one. Its partition function is $Z_j(\gamma)=1$ for all $\gamma$, so \cref{eq:partition} has no solution at $\lambda\ne1$. The bisection routine reaches its iteration limit and leaves the column unchanged. On every graph, the number of columns that reach the limit equals the number of single-neighbor columns, and every column with $s_j\ge2$ converges.

For edges whose endpoints share no neighbors, Jaccard weighting assigns the machine epsilon $\epsilon_{\mathrm{mach}}$, preserving positive probabilities on the support. Equal weights give a uniform column, which remains invariant by \cref{cor:uniform}. Rescaling can act on nodes with at least two neighbors and unequal weights. These account for 53\% of nodes on Cora, 24\% on PubMed, 90\% on ca-AstroPh, and all nodes on MNIST-kNN and \pbmc.

\subsection{Paired comparisons across seeds}
\label{app:stats}

\input{images/t05_paired}

\Cref{tab:paired} pairs methods by seed. We use an exact two-sided Wilcoxon signed-rank test for the four complete ten-seed comparisons. We apply Holm adjustment within two retrospectively defined families. One groups the external-label comparisons on Citeseer and MNIST-kNN. The other groups the two \pbmc metrics. The table also gives adjustments across all four comparisons. With ten nonzero differences and no ties, the smallest attainable two-sided $p$ value rounds to $0.002$ and occurs when all ten differences have the same sign. We summarize the five-seed comparisons on Cora and PubMed, the ca-AstroPh node2vec runs, and the subsampled MNIST PHATE run descriptively.

\section{Reproducibility Artifact}
\label{app:artifact}

Our code is available at \url{https://github.com/Q9gJYx/LambdaLens}. It includes the dataset loaders, embedding and evaluation scripts, and the procedure for deriving the \pbmc pseudo-labels. The repository pins the embedding package revision and records its dependencies in a lock file. The paper source includes the plotted sensitivity values, the degree and layout-displacement data, and the twelve embedding panels in \cref{fig:response_portrait}.

\end{document}

%% file: math_commands.tex
\usepackage{amsmath,amsfonts,bm}

\def\eqref#1{equation~\ref{#1}}

\def\1{\bm{1}}

\DeclareMathAlphabet{\mathsfit}{\encodingdefault}{\sfdefault}{m}{sl}
\SetMathAlphabet{\mathsfit}{bold}{\encodingdefault}{\sfdefault}{bx}{n}

%% file: images/_figstyle.tex
\definecolor{dsCora}    {rgb}{0.84, 0.33, 0.30}  % red
\definecolor{dsPubmed}  {rgb}{0.55, 0.42, 0.69}  % purple
\definecolor{dsCiteseer}{rgb}{0.55, 0.34, 0.29}  % brown
\definecolor{dsAstroph} {rgb}{0.88, 0.55, 0.16}  % orange
\definecolor{dsPbmc}    {rgb}{0.27, 0.46, 0.71}  % blue
\definecolor{dsMnist}   {rgb}{0.36, 0.62, 0.36}  % green

\pgfplotsset{
  /pgfplots/ll plot/.style={
    grid=both,
    grid style={line width=0.2pt, draw=gray!22},
    major grid style={line width=0.2pt, draw=gray!30},
    tick label style={font=\scriptsize},
    label style={font=\scriptsize},
    title style={font=\scriptsize, yshift=-2pt},
    every axis plot/.append style={line width=0.6pt, mark size=1.2pt},
    axis line style={line width=0.4pt, draw=gray!55!black},
    tick style={line width=0.3pt, color=gray!55!black},
  },
}

\tikzset{
  ll cell/.style={
    draw, line width=0.5pt, rounded corners=2pt,
    minimum height=11mm,
    align=center, inner xsep=5pt, inner ysep=4pt,
  },
  ll ours/.style={
    ll cell,
    draw=tabfirst!75!black, fill=tabfirst!45,
    line width=0.6pt,
  },
  ll io/.style={
    ll cell,
    draw=gray!60, fill=gray!12,
    minimum width=20mm,
  },
  ll arrow/.style={
    -{Latex[length=1.8mm,width=1.5mm]},
    line width=0.55pt, gray!50!black,
  },
  ll guide/.style={
    dashed, gray!60, line width=0.4pt,
  },
}

\newcommand{\swatch}[1]{%
  \tikz[baseline=-0.5ex]\draw[color=#1, line width=0.75pt] (0,0) -- (4mm,0);}
\newcommand{\swatchdash}[1]{%
  \tikz[baseline=-0.5ex]\draw[color=#1, line width=0.5pt, dashed] (0,0) -- (4mm,0);}

%% file: images/02_audit_workflow.tex
% Local conditional probabilities in panel (b) are an analytic example.
% Panel (c) reuses the seed-42 Cora portraits and five-seed fidelity summaries.
\begin{figure*}[t]
\centering
\begingroup
\definecolor{llInk}{HTML}{253743}
\definecolor{llSub}{HTML}{445967}
\definecolor{llLine}{HTML}{CAD5DB}
\definecolor{llA}{HTML}{276D80}
\definecolor{llB}{HTML}{C07738}
\definecolor{llC}{HTML}{80649A}
\resizebox{\textwidth}{!}{%
\begin{tikzpicture}[x=1cm,y=1cm,font=\sffamily\small,text=llInk,
  every node/.style={inner sep=0pt,outer sep=0pt},
  head/.style={anchor=west,font=\sffamily\bfseries\large},
  note/.style={anchor=west,font=\sffamily\footnotesize,text=llSub},
  tinylabel/.style={font=\sffamily\footnotesize},
  link/.style={-{Stealth[length=2mm,width=1.6mm]},line width=.8pt,draw=llSub},
  hub/.style={circle,fill=llInk,minimum size=6.2mm,text=white},
  nbr/.style={circle,minimum size=5.3mm,font=\sffamily\bfseries\footnotesize,text=white}]
% The contents of each panel share a baseline and a bottom annotation.
\draw[llLine] (0,4.44) -- (17.5,4.44);
\node[head] at (0,4.78) {(a) Construct affinities};
\node[head] at (4.93,4.78) {(b) Vary the local target};
\node[head] at (11.32,4.78) {(c) Fit paired layouts};
\node[note] at (0.02,4.10) {One stochastic neighborhood};
\node[note] at (4.93,4.10) {The same three neighbors across $\lambda$};
\node[note] at (11.33,4.10) {Cora layouts at seed 42, scores over five seeds};
% A focal node and its three incident probabilities.
\coordinate (j) at (0.94,2.66);
\coordinate (a) at (3.20,3.53);
\coordinate (b) at (3.45,2.64);
\coordinate (c) at (3.12,1.76);
\draw[llA,line width=2.1pt] (j)--(a);
\draw[llB,line width=1.3pt] (j)--(b);
\draw[llC,line width=.7pt] (j)--(c);
\node[hub] at (j) {$j$};
\node[nbr,fill=llA] at (a) {a};
\node[nbr,fill=llB] at (b) {b};
\node[nbr,fill=llC] at (c) {c};
\node[tinylabel,fill=white,inner xsep=2pt] at (2.32,3.19) {$0.60$};
\node[tinylabel,fill=white,inner xsep=2pt] at (2.36,2.65) {$0.30$};
\node[tinylabel,fill=white,inner xsep=2pt] at (2.24,2.12) {$0.10$};
\draw[link] (4.03,2.68)--(4.60,2.68);
% All three glyphs use the same probability scale and neighbor order.
\foreach \x/\lam/\action/\pa/\pb/\pc in
  {5.80/1/Identity/.6/.3/.1,7.71/3/Uniform/.33333/.33333/.33333,9.62/20/Inverted/.09007/.20017/.70976}{
  \begin{scope}[shift={(\x,0)}]
    \node[font=\sffamily\bfseries\small] at (0,3.6) {$\lambda=\lam$};
    \fill[llA] (-.55,2.13) rectangle (-.25,{2.13+1.67*\pa});
    \fill[llB] (-.15,2.13) rectangle (.15,{2.13+1.67*\pb});
    \fill[llC] (.25,2.13) rectangle (.55,{2.13+1.67*\pc});
    \draw[llSub,line width=.4pt] (-.69,2.13)--(.69,2.13);
    \foreach \pos/\name in {-.4/a,0/b,.4/c}
      \node[tinylabel] at (\pos,1.97) {\name};
    \node[font=\sffamily\bfseries\footnotesize] at (0,1.57) {\action};
  \end{scope}
}
\draw[link] (10.53,2.68)--(11.10,2.68);
% Actual Cora layouts, at the two endpoint settings illustrated above.
\foreach \x/\lam/\score in {12.72/1/0.922,15.94/20/0.966}{
  \node at (\x,2.80)
    {\includegraphics[height=1.87cm]{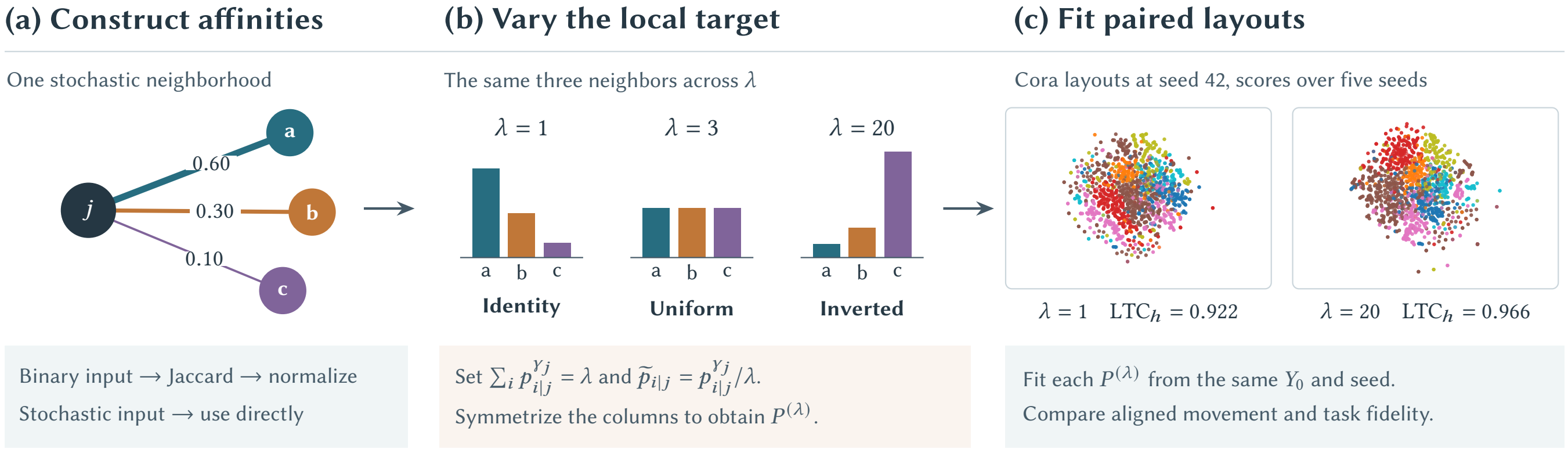}};
  \draw[llLine,rounded corners=2pt] (\x-1.49,1.78) rectangle (\x+1.49,3.81);
  \node[tinylabel] at (\x,1.52) {$\lambda=\lam$\quad$\LTCh=\score$};
}
\fill[llA!7] (0,0) rectangle (4.52,1.14);
\fill[llB!8] (4.88,0) rectangle (10.84,1.14);
\fill[llA!7] (11.23,0) rectangle (17.50,1.14);
\node[note] at (.16,.78) {Binary input $\to$ Jaccard $\to$ normalize};
\node[note] at (.16,.36) {Stochastic input $\to$ use directly};
\node[note] at (5.05,.78) {Set $\sum_i p_{i\mid j}^{\gamma_j}=\lambda$ and $\widetilde p_{i\mid j}=p_{i\mid j}^{\gamma_j}/\lambda$.};
\node[note] at (5.05,.36) {Symmetrize the columns to obtain $P^{(\lambda)}$.};
\node[note] at (11.42,.78) {Fit each $P^{(\lambda)}$ from the same $Y_0$ and seed.};
\node[note] at (11.42,.36) {Compare aligned movement and task fidelity.};
\end{tikzpicture}}
\endgroup
\caption{\textbf{How a change in $\lambda$ reaches the layout.}
In the three-neighbor example (a--b), each color tracks one neighbor as the target changes.
At $\lambda=3$ the neighbors have equal weight. At $\lambda=20$ the least likely neighbor becomes the most likely.
Panel (c) shows two measured Cora layouts with the graph, initialization, optimizer settings, and visual encoding held fixed. The scores combine five-seed means of label trustworthiness and label continuity.
The audit measures geometric change by Procrustes alignment and evaluates fidelity on the same nodes.}
\Description{A node has three neighbors with probabilities 0.60, 0.30, and 0.10. Bars for lambda 1, 3, and 20 show the original, uniform, and reversed order. Two Cora embeddings use seed 42. Five-seed summaries give harmonic label fidelity scores of 0.922 at lambda 1 and 0.966 at lambda 20.}
\label{fig:audit_workflow}
\end{figure*}

%% file: images/t00_datasets.tex
\begin{table*}[t]
  \centering
  \caption{\textbf{Graphs, features, and evaluation labels.} \ours takes the
graph in the third column. Feature-based methods use the available
feature matrix. For graph-only inputs, see \cref{sec:baseline_protocol}.
The \pbmc pseudo-labels come from a spectral partition computed
independently of $\lambda$.}
\label{tab:datasets}
  \small
  \setlength{\tabcolsep}{5pt}
  \resizebox{\textwidth}{!}{%
  \begin{tabular}{l r l l l l}
    \toprule
    Dataset & $n$ & Graph to \ours & Feature space for metrics/baselines
      & Label provenance & Primary metric \\
    \midrule
    Cora & 2,708 & citation adjacency $\rightarrow$ Jaccard stochastic
      & bag-of-words & 7 document classes & $\LTh$ \\
    Citeseer & 3,327 & citation adjacency $\rightarrow$ Jaccard stochastic
      & bag-of-words & 6 document classes & $\LTh$ \\
    PubMed & 19,717 & citation adjacency $\rightarrow$ Jaccard stochastic
      & TF/IDF-like attributes & 3 document classes & $\LTh$ \\
    MNIST-kNN & 70,000 & Euclidean 15-NN $\rightarrow$ Jaccard stochastic
      & 784 pixels & digit label & $\LTh$ \\
    \pbmc & 8,381 & supplied stochastic 30-NN graph
      & adjacency rows (graph-only) & Ward, $k=7$, Laplacian embedding & $\LTh$ \\
    ca-AstroPh & 18,772 & coauthorship adjacency $\rightarrow$ Jaccard stochastic
      & adjacency rows (graph-only) & none & $T$ \\
    \bottomrule
  \end{tabular}%
  }
\end{table*}

%% file: images/01_response_portrait.tex
\begin{figure*}[t]
\newcommand{\responsepanel}[2]{%
  \begin{tikzpicture}[baseline=0pt]
    \node[anchor=south west,inner sep=0pt] (img) at (0,0)
      {\includegraphics[width=0.247\linewidth]{#1}};
    \node[fill=white,fill opacity=0.86,text opacity=1,
          anchor=south east,inner sep=1.5pt,
          font=\fontsize{7}{8}\selectfont\bfseries]
      at (img.south east) {#2};
  \end{tikzpicture}%
}
\centering
\setlength{\tabcolsep}{1pt}
\begin{tabular}{@{}cccc@{}}
  \responsepanel{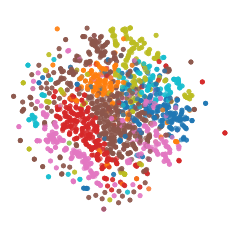}{0.922} &
  \responsepanel{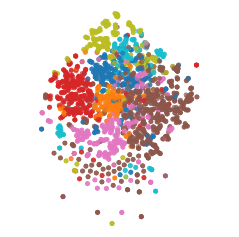}{0.951} &
  \responsepanel{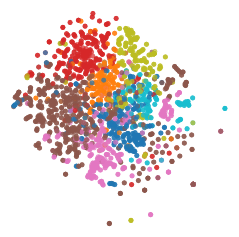}{0.966} &
  \responsepanel{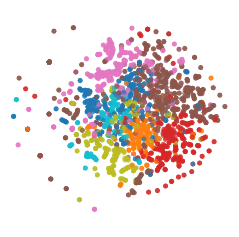}{0.949} \\[1pt]
  \responsepanel{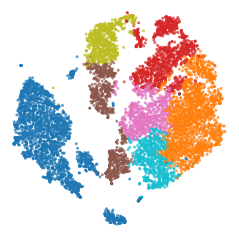}{0.983} &
  \responsepanel{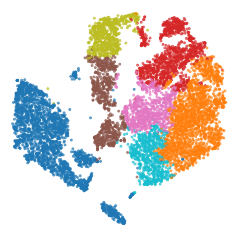}{0.987} &
  \responsepanel{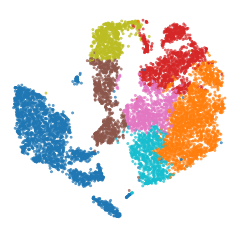}{0.986} &
  \responsepanel{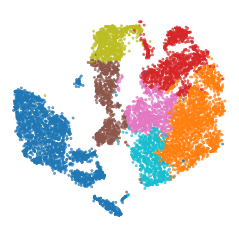}{0.986} \\[1pt]
  \responsepanel{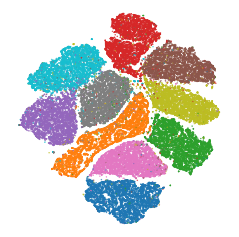}{0.986} &
  \responsepanel{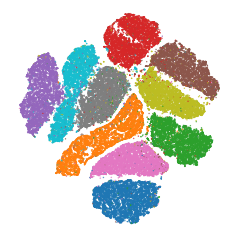}{0.985} &
  \responsepanel{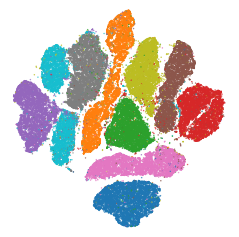}{0.987} &
  \responsepanel{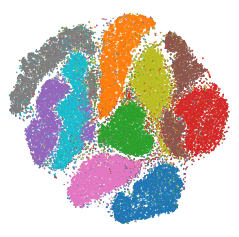}{0.984} \\
\end{tabular}
\caption{\textbf{Layout and label fidelity as $\lambda$ increases.}
Rows show Cora, \pbmc, and MNIST-kNN. From left to right,
$\lambda=1,5,20,80$. Panels within a row share the same graph,
seed~42, adjacency-SVD initialization, optimizer settings, and color
encoding. On Cora and MNIST-kNN, the inset score is the harmonic mean
of $\LTh$ and $\LCh$ after averaging each over seeds 42--46.
On \pbmc, the inset gives $\LTCh$ for the displayed seed on 5,000 nodes.
Colors indicate document classes on
Cora, spectral pseudo-labels on \pbmc, and digits on MNIST.
Cora continues to move across settings, \pbmc settles after $5$,
and the MNIST layout changes little.}
\Description{A three-by-four grid of two-dimensional embeddings for
Cora, PBMC-8k, and MNIST-kNN at lambda values 1, 5, 20, and 80. Cora
changes across several columns, PBMC changes mostly between the first
two, and MNIST remains visually stable.}
\label{fig:response_portrait}
\end{figure*}
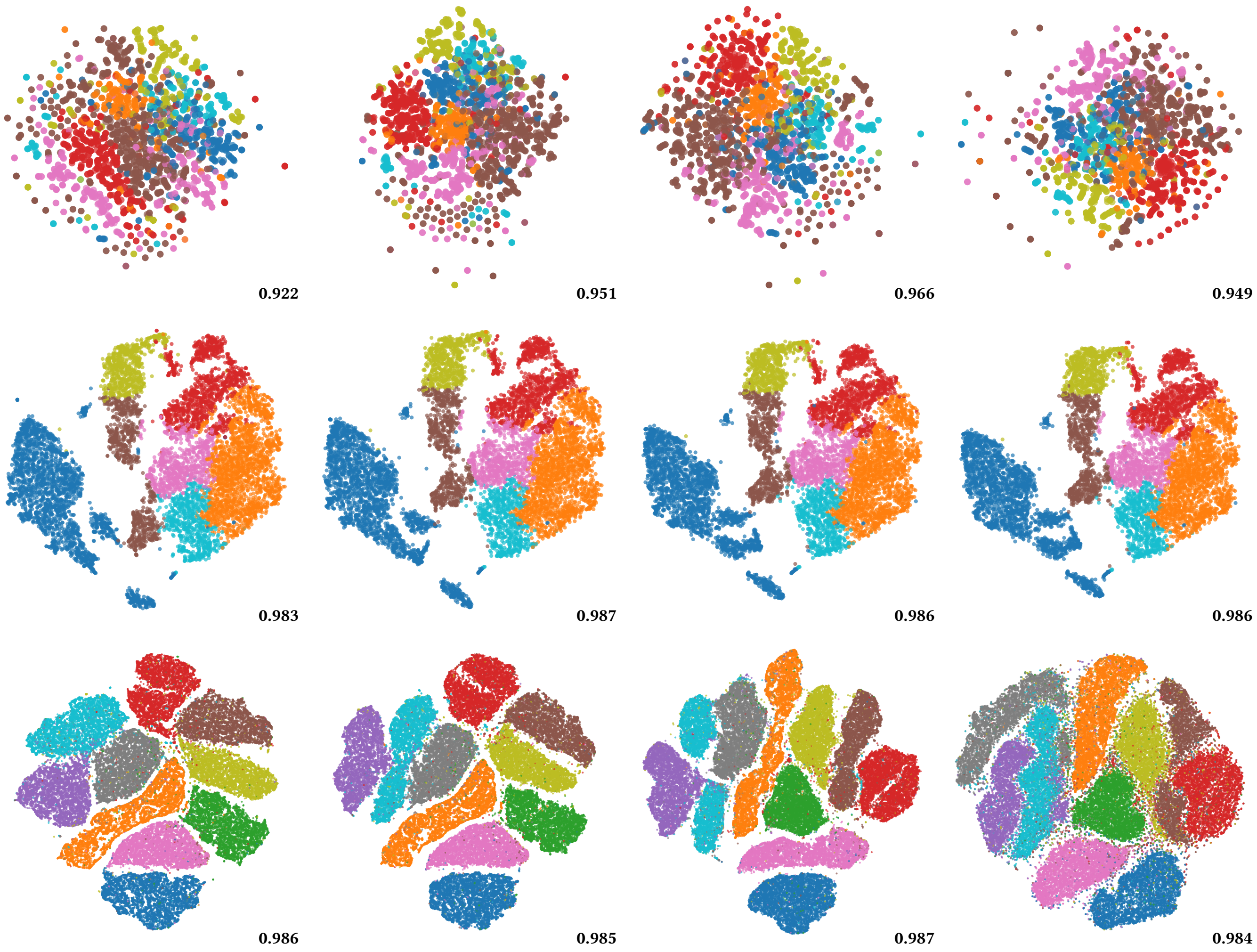

%% file: images/04_diagnostic.tex
\begin{figure}[t]
  \centering
  \begin{tikzpicture}
    \begin{axis}[
        ll plot,
        width=\columnwidth,
        height=0.84\columnwidth,
        xlabel={$\mathrm{CV}(d)$},
        ylabel={mean pairwise $\Proc$},
        xmin=0.15,xmax=1.82,
        ymin=0.03,ymax=0.93,
        xtick={0.3,0.6,0.9,1.2,1.5,1.8},
        ytick={0.1,0.3,0.5,0.7,0.9},
        clip=false,
      ]
      \addplot[only marks,mark=*,mark size=2.5pt,color=dsCora] coordinates {(1.3411,0.8538)};
      \addplot[only marks,mark=*,mark size=2.5pt,color=dsCiteseer] coordinates {(1.2207,0.7599)};
      \addplot[only marks,mark=*,mark size=2.5pt,color=dsPubmed] coordinates {(1.6528,0.1932)};
      \addplot[only marks,mark=*,mark size=2.5pt,color=dsAstroph] coordinates {(1.4485,0.5615)};
      \addplot[only marks,mark=*,mark size=2.5pt,color=dsPbmc] coordinates {(0.7513,0.1103)};
      \addplot[only marks,mark=*,mark size=2.5pt,color=dsMnist] coordinates {(0.2997,0.0954)};

      \node[anchor=west,font=\scriptsize,text=dsCora!70!black,xshift=3pt]
        at (axis cs:1.3411,0.8538) {Cora};
      \node[anchor=east,font=\scriptsize,text=dsCiteseer!70!black,xshift=-3pt]
        at (axis cs:1.2207,0.7599) {Citeseer};
      \node[anchor=west,font=\scriptsize,text=dsAstroph!70!black,xshift=3pt]
        at (axis cs:1.4485,0.5615) {ca-AstroPh};
      \node[anchor=south,font=\scriptsize,text=dsPubmed!70!black,yshift=3pt]
        at (axis cs:1.6528,0.1932) {PubMed};
      \node[anchor=south,font=\scriptsize,text=dsPbmc!70!black,yshift=3pt]
        at (axis cs:0.7513,0.1103) {\pbmc};
      \node[anchor=south,font=\scriptsize,text=dsMnist!70!black,yshift=3pt]
        at (axis cs:0.2997,0.0954) {MNIST-kNN};

      \draw[-{Latex[length=1.4mm]},gray!65,line width=.45pt]
        (axis cs:1.58,0.69) -- (axis cs:1.64,0.23);
      \node[anchor=east,align=right,font=\scriptsize,text=gray!70!black]
        at (axis cs:1.57,0.70) {high degree dispersion,\\low response};
    \end{axis}
  \end{tikzpicture}
  \caption{\textbf{PubMed has the largest relative spread in degree but little layout movement.}
Each point represents one graph. The horizontal axis shows the degree
coefficient of variation $\mathrm{CV}(d)$. The vertical axis shows
normalized Procrustes distance averaged over all pairs of layouts on
the 16-point grid at seed~42. Cora and Citeseer move farther than
PubMed despite having lower degree dispersion.}
\Description{A scatter plot of degree coefficient of variation
  against mean pairwise aligned embedding movement for six datasets.
  PubMed has the highest degree dispersion but low movement,
  contradicting a one-statistic response rule.}
  \label{fig:cv_procrustes}
\end{figure}
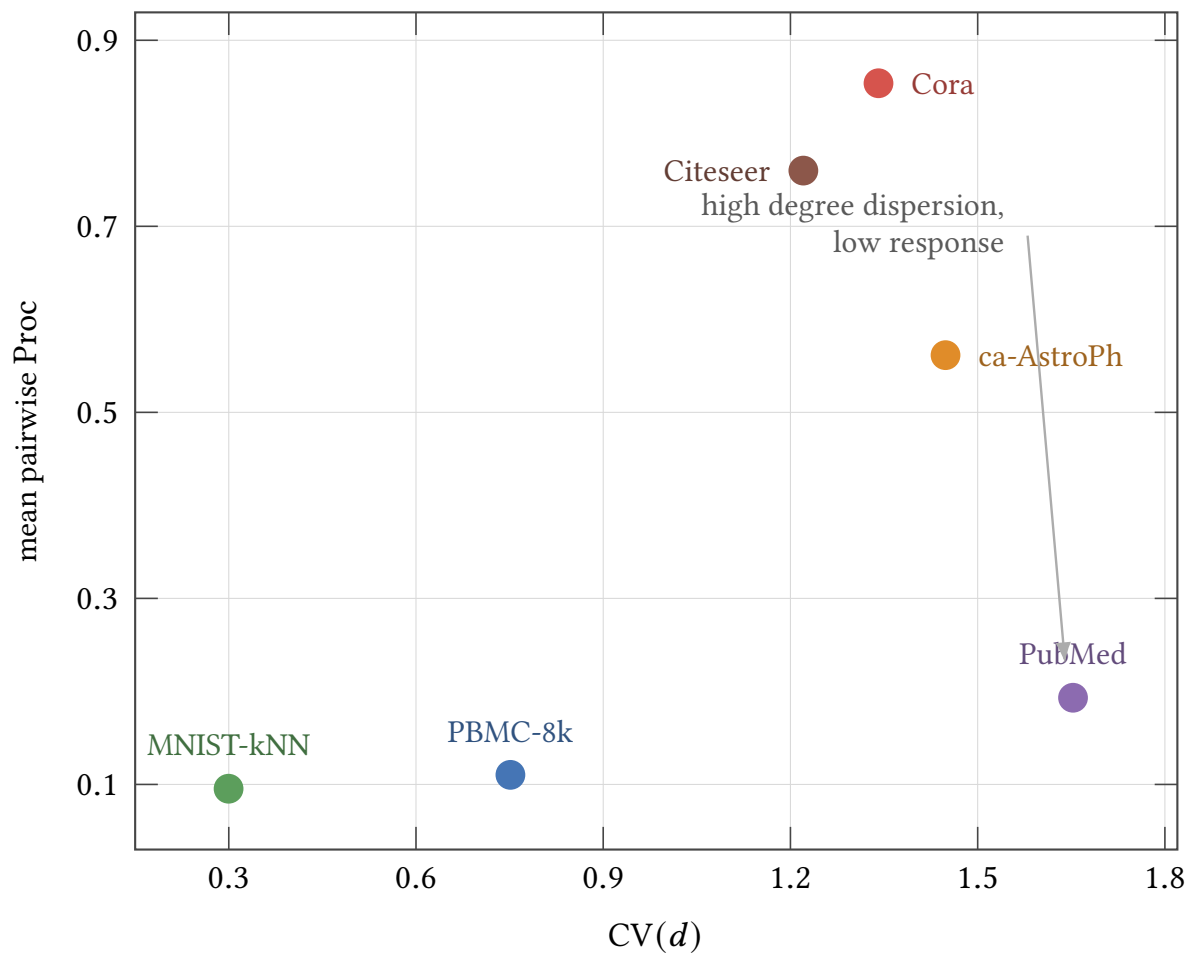

%% file: images/t01_probe_search.tex
\begin{table}[t]
  \centering
  \small
  \caption{\textbf{Best label-fidelity scores on the four- and 16-point grids.}
Both searches use seed~42 with the same preprocessing, initial
coordinates, labels, and metric. Loss is the percentage decrease
relative to the best dense-grid score, computed from the displayed values.}
\label{tab:probe_search}
  \setlength{\tabcolsep}{4pt}
  \begin{tabular}{l c c c c}
    \toprule
    Dataset & $\widehat\lambda_4$ & $\widehat\lambda_{16}$
      & $\LTCh^{(4)}/\LTCh^{(16)}$ & Loss \\
    \midrule
    Cora      & 20 & 7 & $0.9664/0.9668$ & $0.04\%$ \\
    Citeseer  & 1 & 3 & $0.8536/0.8668$ & $1.52\%$ \\
    PubMed    & 5 & 5 & $0.9543/0.9543$ & $0.00\%$ \\
    MNIST-kNN & 1 & 30 & $0.9868/0.9881$ & $0.13\%$ \\
    \bottomrule
  \end{tabular}
\end{table}

%% file: images/t02_comparison.tex
\begin{table*}[t]
  \centering
  \caption{\textbf{Embedding fidelity and wall-clock time in seconds.}
  Scores are raw Label-Trustworthiness ($\LTh$) on labeled graphs and
  trustworthiness ($T$) on ca-AstroPh. We report the mean and sample standard
  deviation over ten seeds on Citeseer, MNIST-kNN, and \pbmc and five elsewhere,
  with the exceptions listed below. For \ours, the four probes select $\lambda=20$
  on Cora, $1$ on Citeseer, and $5$ on PubMed. The benchmark uses $20$ on
  MNIST-kNN and $10$ on both \pbmc and ca-AstroPh.
  Baseline settings appear in \cref{sec:baseline_protocol}.}
  \label{tab:comparison}
  \scriptsize
  \setlength{\tabcolsep}{3pt}
  \begin{tabular}{l cc cc cc cc cc cc}
    \toprule
    & \multicolumn{2}{c}{Cora}
    & \multicolumn{2}{c}{Citeseer}
    & \multicolumn{2}{c}{PubMed}
    & \multicolumn{2}{c}{MNIST-kNN}
    & \multicolumn{2}{c}{\pbmc}
    & \multicolumn{2}{c}{ca-AstroPh} \\
    \cmidrule(lr){2-3} \cmidrule(lr){4-5} \cmidrule(lr){6-7}
    \cmidrule(lr){8-9} \cmidrule(lr){10-11} \cmidrule(lr){12-13}
    Method
      & $\LTh\uparrow$ & $s\downarrow$
      & $\LTh\uparrow$ & $s\downarrow$
      & $\LTh\uparrow$ & $s\downarrow$
      & $\LTh\uparrow$ & $s\downarrow$
      & $\LTh\uparrow$ & $s\downarrow$
      & $T\uparrow$    & $s\downarrow$ \\
    \midrule
    UMAP~\cite{mcinnes2018umap}
      & $0.804$\s{0.004} & $10$
      & $0.617$\s{0.005} & $34$
      & $0.880$\s{0.016} & $41$
      & \best{1.000}\s{0.000} & $162$
      & $0.981$\s{0.011} & $25$
      & $0.594$\s{0.003} & $125$ \\
    openTSNE~\cite{policar2024opentsne}
      & $0.765$\s{0.004} & $12$
      & $0.766$\s{0.010} & $18$
      & $0.883$\s{0.009} & $87$
      & $0.997$\s{0.004} & $262$
      & $0.723$\s{0.004} & $26$
      & $0.620$\s{0.002} & $162$ \\
    node2vec+UMAP~\cite{grover2016node2vec}
      & $0.915$\s{0.007} & $420$
      & $0.746$\s{0.013} & $432$
      & $0.857$\s{0.018} & $354$
      & -- & --
      & $0.966$\s{0.015} & $407$
      & $0.621$\s{0.003}$^{\dagger}$ & $4083^{\dagger}$ \\
    PHATE~\cite{moon2019phate}
      & $0.762$\s{0.004} & $4$
      & \best{0.855}\s{0.006} & $8$
      & $0.895$\s{0.008} & $18$
      & -- & $102$$^{\ddagger}$
      & $0.922$\s{0.006} & $12$
      & $0.509$\s{0.002}$^{\ddagger}$ & $64$ \\
    \midrule
    \ours ($\lambda$: see caption)
      & \best{0.932}\s{0.013} & $11$
      & $0.754$\s{0.012} & $8$
      & \best{0.905}\s{0.005} & $46$
      & $0.991$\s{0.007} & $163$
      & \best{0.989}\s{0.003} & $22$
      & \best{0.622}\s{0.003} & $30$ \\
    \bottomrule
  \end{tabular}\\[2pt]
  \parbox{\linewidth}{\scriptsize\textit{Notes.}
  Bold marks the highest reported fidelity on each graph. On \pbmc, $\LTh$ measures
  agreement with a $\lambda$-independent spectral partition. A dash marks an
  unavailable score or one computed at a different evaluation size.
  Node2vec+UMAP uses five seeds on Citeseer and \pbmc.
  $\dagger$ The ca-AstroPh score covers five seeds. The time is from the one run
  with a fixed thread count. $\ddagger$ PHATE uses 10,000 nodes on MNIST and
  3,000 on ca-AstroPh. Times depend on the implementation, preprocessing, and hardware.}
\end{table*}

%% file: images/03_sensitivity.tex
%% F3 — Sensitivity 2×2 (perplexity / n_iter / k_kNN / alpha) via
%% pgfplots groupplots from per-(panel, dataset) CSVs in images/data/.
%% N=3 seeds. Single-column figure; default \ours setting per panel
%% marked by a dashed vertical guide. Single-column placement chosen
%% because each panel has only 5-6 sweep points; full-width would be
%% wasteful given data sparsity. Dataset colors: shared palette
%% (images/_figstyle.tex), color encodes identity.

\begin{figure}[t]
  \centering
  \begin{tikzpicture}
    \begin{groupplot}[
        ll plot,
        group style={
          group size=2 by 2,
          horizontal sep=24pt,
          vertical sep=46pt,
          ylabels at=edge left,
          xlabels at=edge bottom,
        },
        width=0.57\columnwidth,
        height=0.54\columnwidth,
      ]

      %% ============== (a) perplexity ==============
      \nextgroupplot[
        title={(a) perplexity},
        xlabel={perplexity},
        ylabel={$\LTCh$},
        xmode=log, log basis x=10,
        xmin=8, xmax=160,
        xtick={10,30,80,120}, xticklabels={10,30,80,120},
        ymin=0.55, ymax=1.02,
        ytick={0.6,0.8,1.0},
      ]
      \addplot[mark=*, color=dsCora] table[col sep=comma, x=value, y=mean]
        {images/data/sens_perplexity_cora.csv};
      \addplot[mark=*, color=dsMnist] table[col sep=comma, x=value, y=mean]
        {images/data/sens_perplexity_mnist_knn.csv};
      \draw[ll guide] (axis cs:30,0.55) -- (axis cs:30,1.02);

      %% ============== (b) n_iter ==============
      \nextgroupplot[
        title={(b) $n_{\mathrm{iter}}$},
        xlabel={$n_{\mathrm{iter}}$},
        xmode=log, log basis x=10,
        xmin=400, xmax=10000,
        xtick={500,1000,2000,4000,8000},
        xticklabels={.5k,1k,2k,4k,8k},
        ymin=0.96, ymax=1.0,
        ytick={0.96,0.98,1.0},
      ]
      \addplot[mark=*, color=dsCora] table[col sep=comma, x=value, y=mean]
        {images/data/sens_niter_cora.csv};
      \addplot[mark=*, color=dsPbmc] table[col sep=comma, x=value, y=mean]
        {images/data/sens_niter_pbmc.csv};
      \addplot[mark=*, color=dsMnist] table[col sep=comma, x=value, y=mean]
        {images/data/sens_niter_mnist_knn.csv};
      \draw[ll guide] (axis cs:1000,0.96) -- (axis cs:1000,1.0);

      %% ============== (c) k_kNN ==============
      \nextgroupplot[
        title={(c) $k_{\mathrm{kNN}}$ (MNIST-kNN only)},
        xlabel={$k_{\mathrm{kNN}}$},
        ylabel={$\LTCh$},
        xmin=3, xmax=55,
        xtick={5,15,30,50},
        ymin=0.96, ymax=1.0,
        ytick={0.96,0.98,1.0},
      ]
      \addplot[mark=*, color=dsMnist] table[col sep=comma, x=value, y=mean]
        {images/data/sens_k_knn_mnist_knn.csv};
      \draw[ll guide] (axis cs:15,0.96) -- (axis cs:15,1.0);

      %% ============== (d) alpha ==============
      \nextgroupplot[
        title={(d) $\alpha$ (early-exaggeration)},
        xlabel={$\alpha$},
        xmode=log, log basis x=10,
        xmin=0.7, xmax=70,
        xtick={1,5,12,50}, xticklabels={1,5,12,50},
        ymin=0.86, ymax=1.0,
        ytick={0.88,0.92,0.96,1.0},
      ]
      \addplot[mark=*, color=dsCora] table[col sep=comma, x=value, y=mean]
        {images/data/sens_alpha_cora.csv};
      \addplot[mark=*, color=dsPbmc] table[col sep=comma, x=value, y=mean]
        {images/data/sens_alpha_pbmc.csv};
      \addplot[mark=*, color=dsMnist] table[col sep=comma, x=value, y=mean]
        {images/data/sens_alpha_mnist_knn.csv};
      \draw[ll guide] (axis cs:12,0.86) -- (axis cs:12,1.0);
    \end{groupplot}

    %% Shared legend above the top row, centered across the group bbox
    \node[anchor=south, font=\scriptsize, yshift=4pt]
      at ($(group c1r1.north east)!0.5!(group c2r1.north west)+(0,12pt)$)
      {\swatch{dsCora}\,Cora\quad
       \swatch{dsPbmc}\,\pbmc\quad
       \swatch{dsMnist}\,MNIST-kNN\quad
       \swatchdash{gray!60}\,default};
  \end{tikzpicture}
  \caption{\textbf{Sensitivity of label fidelity to graph construction and optimization.}
Curves report mean $\LTCh$ over $N{=}3$ seeds, varying one setting at a
time. Dashed lines indicate defaults. Panel (a) uses openTSNE to
construct input affinities at different perplexities. \pbmc already
has a weighted graph. Panel (c) varies $k$ on a 10,000-image MNIST
subset using cosine distance. Fidelity is near its plateau at 1,000
iterations, and $k=15$ and $30$ give similar scores. The best observed
early exaggeration is $\alpha=12$ on Cora and \pbmc and $20$ on MNIST.}
\Description{Four line charts show label-fidelity sensitivity to
  upstream affinity perplexity, iteration count, MNIST neighborhood
  size, and early-exaggeration alpha for Cora, PBMC-8k, and MNIST-kNN
  where applicable.}
  \label{fig:sensitivity}
\end{figure}
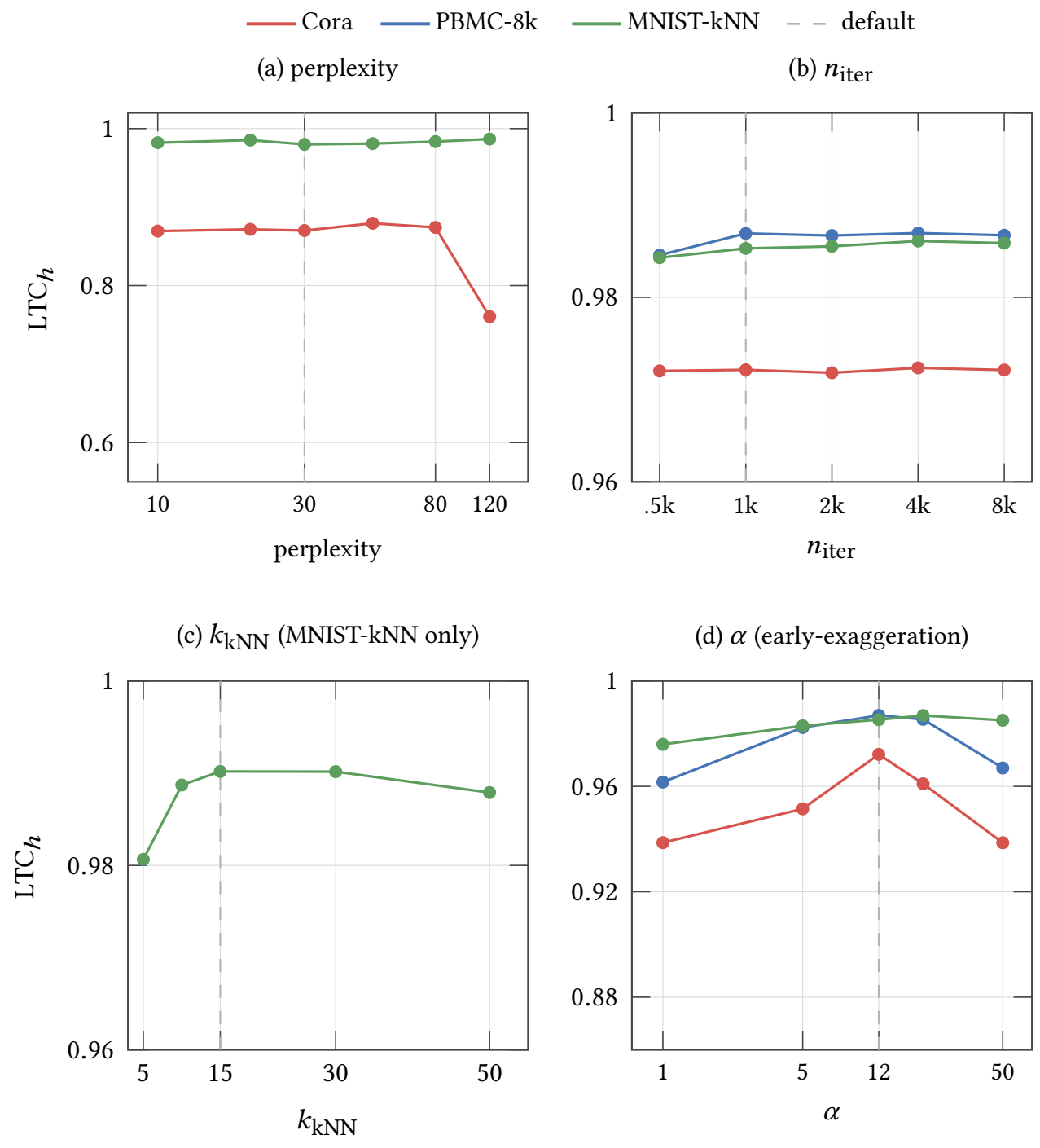

%% file: images/t03_ablation.tex
\begin{table}[t]
  \centering
  \scriptsize
  \setlength{\tabcolsep}{4pt}
  \caption{\textbf{Effect of each model change on raw $\LTh$.}
  Entries are means and sample standard deviations over seeds 42--46 on a
  Linux x86-64 host. The full configuration uses $\lambda=20$ on Cora
  and $\lambda=5$ on \pbmc. The two Cora runs at $20$ use the same settings.
  The stochastic \pbmc graph bypasses Jaccard weighting. The Cora benchmark
  in \cref{tab:comparison} uses the same full configuration on an M3 Pro
  (\cref{app:ablation}).}
  \label{tab:ablation}
  \begin{tabular}{l c c}
    \toprule
    Variant & Cora $\LTh$ & \pbmc $\LTh$ \\
    \midrule
    \rowcolor{gray!15} \ours full
      & $0.935$\s{0.019} & $0.987$\s{0.005} \\
    \quad no adjacency-SVD init
      & $0.912$\s{0.021} & $0.981$\s{0.009} \\
    \quad no Jaccard weighting
      & $0.940$\s{0.003} & -- \\
    \quad $\lambda=1$ (identity)
      & $0.855$\s{0.020} & $0.977$\s{0.004} \\
    \quad $\lambda=20$
      & $0.935$\s{0.019} & $0.988$\s{0.005} \\
    \bottomrule
  \end{tabular}
\end{table}

%% file: images/t04_support.tex
\begin{table}[t]
  \centering
  \scriptsize
  \setlength{\tabcolsep}{4pt}
  \caption{\textbf{How many nodes can change their neighbor weights?}
  The first three count columns group nodes by support size $s_j$ after
  self-loop removal, Jaccard weighting where used, and normalization
  of nonempty columns. Among nodes with $s_j\ge2$, the last two columns count
  uniform neighborhoods and those with $s_j<\lambda$ at the benchmark setting.
  These groups can overlap. Uniform weights stay equal, while unequal
  weights reverse order when $2\le s_j<\lambda$. Citeseer uses the identity
  setting $\lambda=1$ (\cref{prop:phases,cor:uniform,tab:comparison}).}
  \label{tab:support}
  \begin{tabular}{l r r r r r r}
    \toprule
    Graph & $\lambda$ & $s_j=0$ & $s_j=1$ & $s_j\ge2$
      & uniform & $2\le s_j<\lambda$ \\
    \midrule
    Cora       & 20 &  0 &   485 &  2{,}223 &    793 &  2{,}199 \\
    Citeseer   &  1 & 48 & 1{,}331 &  1{,}948 &    985 &        0 \\
    PubMed     &  5 &  0 & 9{,}094 & 10{,}623 & 5{,}836 &  5{,}855 \\
    MNIST-kNN  & 20 &  0 &     0 & 70{,}000 &      0 & 30{,}472 \\
    \pbmc      & 10 &  0 &     0 &  8{,}381 &      0 &        0 \\
    ca-AstroPh & 10 &  1 & 1{,}282 & 17{,}489 &    538 &  8{,}158 \\
    \bottomrule
  \end{tabular}
\end{table}

%% file: images/t05_paired.tex
\begin{table*}[!b]
  \centering
  \scriptsize
  \setlength{\tabcolsep}{4pt}
  \caption{\textbf{Paired differences in fidelity across ten seeds.}
  Each difference is the score for \ours minus the comparator's score for the
  same seed (42--51). The table reports exact two-sided Wilcoxon signed-rank
  $p$ values and Holm-adjusted values for two families defined after analysis,
  external labels (E1, E2) and \pbmc (P1, P2). All differences are nonzero
  with distinct absolute values within each comparison. Across all four tests
  in row order, Holm gives $0.008$, $0.008$, $0.049$, $0.008$ and Bonferroni
  gives $0.008$, $0.008$, $0.195$, $0.008$. The exact Hodges--Lehmann intervals
  have $95.1\%$ coverage. P1 and P2 use the same UMAP runs with one shared
  initialization across seeds.}
  \label{tab:paired}
  \begin{tabular}{l l c c c c c}
    \toprule
    & Comparison & \ours higher & Mean diff.\ [95\% $t$] & HL [95\% exact]
      & $p$ & Holm \\
    \midrule
    E1 & Citeseer $\LTh$ vs PHATE   & 0/10  & $-0.101$ [$-0.110$, $-0.093$] & $-0.102$ [$-0.110$, $-0.093$] & $0.002$ & $0.004$ \\
    E2 & MNIST-kNN $\LTh$ vs UMAP   & 0/10  & $-0.009$ [$-0.014$, $-0.004$] & $-0.008$ [$-0.015$, $-0.004$] & $0.002$ & $0.004$ \\
    \midrule
    P1 & \pbmc pseudo-label $\LTh$ vs UMAP & 9/10  & $+0.008$ [$-0.001$, $+0.016$] & $+0.004$ [$+0.003$, $+0.019$] & $0.049$ & $0.049$ \\
    P2 & \pbmc graph $T$ vs UMAP    & 10/10 & $+0.023$ [$+0.021$, $+0.025$] & $+0.024$ [$+0.021$, $+0.024$] & $0.002$ & $0.004$ \\
    \bottomrule
  \end{tabular}
\end{table*}